\documentclass[10pt, conference, anonymous, compsoc]{IEEEtran}
\usepackage[dvipsnames]{xcolor}

\usepackage{amsfonts}
\usepackage{amssymb}

\usepackage{amsmath}
\ifCLASSOPTIONcompsoc
  \usepackage[nocompress]{cite}
\else
  \usepackage{cite}
\fi

\ifCLASSOPTIONcompsoc
 \usepackage[pdftex]{graphicx}
 \graphicspath{{fig/}}
\else
 \usepackage[dvips]{graphicx}
 \graphicspath{{fig/}}
\fi
\usepackage{graphicx}
\usepackage{textcomp}  
\usepackage{scalerel}  

\usepackage[skip=1pt]{caption}
\usepackage[ruled,vlined,linesnumbered,noresetcount]{algorithm2e}
\usepackage{amssymb}
\usepackage{amsmath}
\usepackage{bm}
\usepackage{array}
\usepackage{multirow}
\def \facepalm{\includegraphics[scale=0.05]{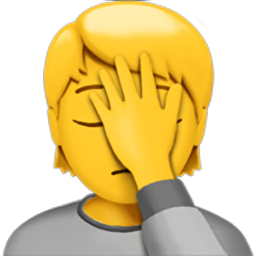}}
\def \finger{\includegraphics[scale=0.05]{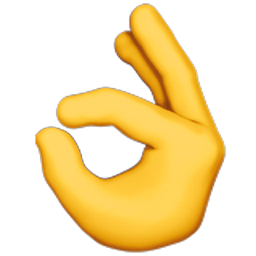}}
\def \tumbsup{\includegraphics[scale=0.05]{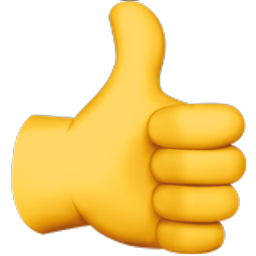}}
\def \positive{\includegraphics[scale=0.05]{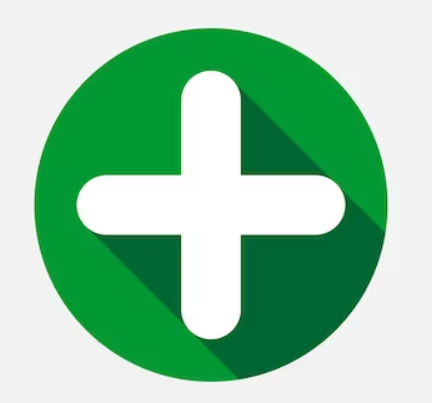}}
\def \negative{\includegraphics[scale=0.05]{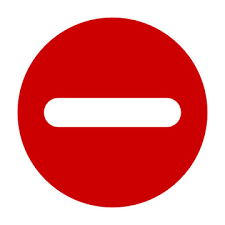}}
\def \neutral{\includegraphics[scale=0.05]{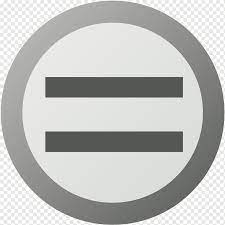}}
\usepackage{csquotes}

\ifCLASSOPTIONcompsoc
 \usepackage[caption=false,font=footnotesize,labelfont=sf,textfont=sf]{subfig}
\else
 \usepackage[caption=false,font=footnotesize]{subfig}
\fi

\usepackage{fixltx2e}

\begin{document}



\title{Leveraging ChatGPT As Text Annotation Tool For Sentiment Analysis}

\author{
    \IEEEauthorblockN{
        Mohammad~Belal\IEEEauthorrefmark{1}
        James She\IEEEauthorrefmark{1}, 
         and Simon~Wong\IEEEauthorrefmark{4} 
    }
    \IEEEauthorblockA{
        \IEEEauthorrefmark{1}College of Science and Engineering, Hamad Bin Khalifa University, Qatar\\
        \IEEEauthorrefmark{4}Department of Electronic and Computer Engineering, HKUST, Hong Kong\\
    }
    \IEEEauthorblockA{
        Email: \IEEEauthorrefmark{1}\{mobe50543,pshe\}@hbku.edu.qa,\IEEEauthorrefmark{4}tywongbf@connect.ust.hk
    }
 }

\maketitle

\IEEEtitleabstractindextext{%
\begin{abstract}
Sentiment analysis is a well-known natural language processing task that involves identifying the emotional tone or polarity of a given piece of text. With the growth of social media and other online platforms, sentiment analysis has become increasingly crucial for businesses and organizations seeking to monitor and comprehend customer feedback as well as opinions. Supervised learning algorithms have been popularly employed for this task, but they require human-annotated text to create the classifier. To overcome this challenge, lexicon-based tools have been used. A drawback of lexicon-based algorithms is their reliance on pre-defined sentiment lexicons, which may not capture the full range of sentiments in natural language. ChatGPT is a new product of OpenAI and has emerged as the most popular AI product. It can answer questions on various topics and tasks. This study explores the use of ChatGPT as a tool for data labeling for different sentiment analysis tasks. It is evaluated on two distinct sentiment analysis datasets with varying purposes.
The results demonstrate that ChatGPT outperforms other lexicon-based unsupervised methods with significant improvements in overall accuracy. Specifically, compared to the best-performing lexical-based algorithms, ChatGPT achieves a remarkable increase in accuracy of 20\%  for the tweets dataset and approximately 25\%  for the Amazon reviews dataset. These findings highlight the exceptional performance of ChatGPT in sentiment analysis tasks, surpassing existing lexicon-based approaches by a significant margin. The evidence suggests it can be used for annotation on different sentiment analysis events and tasks.
\end{abstract}

}

\maketitle

\IEEEdisplaynontitleabstractindextext

\IEEEpeerreviewmaketitle

\section{Introduction}\label{sec:introduction}
Social media platforms such as Twitter, Facebook, Youtube, and Tiktok have become essential tools for users to share their opinions and interact with others online. Twitter, in particular, is a popular micro-blogging platform where users post short but informative content such as text, emojis, and hashtags. These components provide valuable information for sentiment analysis and opinion mining to evaluate users' polarity towards specific topics or events.
Sentiment analysis requires data labeling for the training of the classifiers. These data labels should be of high quality so that the classifier can be closer to the ground truth. One popular method to get the gold standard labeled dataset is using crowd workers for their annotations, such as Amazon Mechanical Turk. The annotation is a time taking and costly process which is sometimes a hindrance for researchers. Moreover, the quality of the crowd workers is also decreased \cite{doi:10.1177/1948550619875149}.

Recently there has been an increase in the popularity of large language models for different NLP tasks. These language models have zero-shot ability to perform different tasks \cite{zhang2022opt}\cite{chowdhery2022palm}. ChatGPT is a new chatbot by OpenAI trained on GPT 3.5 and using RLHF technique to align it with humans \cite{rlhf}. It is a powerful tool that can write code, poetry and can have meaningful conversations with people. It has shown its ability on various NLP tasks, including arithmetic reasoning, text classification, and question answering\cite{qin2023chatgpt}. 
ChatGPT could also be used as a text annotator and shows its superiority among crowd workers for data labeling \cite{gilardi2023chatgpt}.

\begin{table}[h!]
\caption{ChatGPT vs different lexicon-based algorithms}
\begin{center}
\begin{tabular}{ | m{11em} | m{0.8cm}|m{0.9cm}|m{0.9cm}|m{0.7cm}|} 
 \hline
 Text & VADER & TextBlob &  ChatGPT & Ground Truth\\ [0.5ex] 
 \hline\hline

  \#MUNLIV Fred!!!! \facepalm \facepalm  & \neutral & \neutral & \negative & \negative
  \\ 
  \hline
  Get in there!!!!! \#GGMU \finger & \positive & \neutral & \positive & \positive
  \\
  \hline
   Gini’s turned up, now if the rest of the team could, that’d be great! \tumbsup  \#MUNLIV & \positive & \positive & \negative & \negative
    \\
   \hline
   This Apex seems to be a fine unit, except the sound quality on the DVD side is quite poor. Quality otherwise has been good, unlike some of the previous reviews. Have they improved these, or did I just get a good one?
    & \positive & \positive & \negative & \negative
    \\
    \hline

\end{tabular}
\end{center}
\label{table:tweet sentiment}
\end{table}
Lexicon-based methods are commonly used to determine sentiment in texts due to their simplicity and speed compared to supervised learning approaches. VADER, SentiStrength, and TextBlob are popular lexicon-based algorithms. Among these, VADER, a rule-based algorithm, is efficient and outperforms other algorithms on social media data, especially Twitter. It also considers emojis when calculating sentiment score. However, TextBlob's algorithm does not consider emojis when determining sentiment. 
The limitation of lexicon-based algorithms for sentiment analysis is their inability to handle sarcasm, irony, and other forms of figurative language. These algorithms rely on pre-defined lists of words and their associated sentiment scores, which can lead to inaccuracies when processing text that contains words with multiple meanings or words used in a non-literal sense. In addition, these algorithms often struggle with understanding the context and tone of the text, leading to misclassification of the sentiment.
For instance, example 3 in table \ref{table:tweet sentiment} uses sarcastic language to criticize the tweet. Both lexical-based tools could not identify the sarcasm and output the sentiment as positive, whereas ChatGPT understands the sarcasm and explains why it is negative. 
 Our experiments show that the ChatGPT could be used as a tool for sentiment analysis, and it has the ability to understand emojis, sarcasm, and irony. The main findings of this evaluation are as follows
 \begin{enumerate}
    \item ChatGPT impressive performance on zero-shot sentiment analysis
    \item Handles emojis and sarcasm into the sentiment calculations
    \item More than 94\% accurate on the long form of sentiment reviews
\end{enumerate}

Sections 1-6 cover the introduction, related works, problem formulation, experimental setup and results, discussion, and conclusion.

\section{Related Works}\label{sec:related_works}
This section reviews prior work related to large language models and sentiment analysis techniques.

\textbf{Large Language Models:}
There has been an increase in the development of large language models. Specifically after the release of GPT-3 \cite{gpt3}. GPT-3 has been trained auto regressively on a large amount of internet data and has 175 billion parameters. These large language models \cite{gpt3} \cite{chowdhery2022palm} \cite{zhang2022opt} can perform a variety of tasks without much training. These models could perform much better on finetuning on a smaller dataset. These models are few-shot learners. OpenAI has recently launched ChatGPT, a conversational artificial intelligence system that has been fine-tuned from GPT-3.5 using reinforcement learning from human feedback (RLHF) \cite{rlhf}. Large language models can capture context, sarcasm, and other subtleties in sentiment expression when finetuned for that specific task.

\textbf{Sentiment Analysis:}
The objective of sentiment analysis is to recognize the viewpoints, feelings, and emotions conveyed in textual data, including customer feedback, social media updates, and news stories \cite{sentiment}. There have been multiple proposals for evaluating the sentiment of text, including the utilization of machine learning methods that can classify sentiment automatically. It has been demonstrated that both supervised and unsupervised learning approaches can be successful in this regard.Unlike traditional lexicon-based approaches, which rely on predefined sentiment lexicons, large language models can capture context, sarcasm, and other subtleties in sentiment expression.
Yue et al.~\cite{surveySentimentAnalysis} conducted a study comparing the performance of supervised and unsupervised machine learning techniques for sentiment analysis. Their findings revealed that supervised methods generally exhibit superior accuracy than unsupervised approaches like lexicon-based algorithms. Nonetheless, acquiring sufficient labeled training data for supervised methods can be costly and time-consuming.

\section{Problem Formulation}\label{sec:Problem Formulation}
Sentiment analysis is vital in natural language processing to identify the text's emotional tone or polarity. As social media and online platforms gain prominence, businesses and organizations encounter the need to monitor and comprehend customer feedback and opinions efficiently. Although traditional supervised learning algorithms have been commonly employed for sentiment analysis, their dependence on labeled training data poses cost and time constraints.

Recently, there has been a notable increase in the fascination surrounding language models such as ChatGPT, developed by OpenAI, for various natural language processing (NLP) tasks. ChatGPT, built on the powerful GPT-3.5 architecture and trained on extensive text data, has demonstrated exceptional proficiency in generating human-like responses and comprehending diverse subjects comprehensively. With such impressive capabilities, it becomes imperative to explore the potential utility of ChatGPT as a valuable instrument for text annotation.
This research aims to investigate the efficacy of ChatGPT as a text annotation tool for sentiment analysis tasks. Our primary objective is to assess the ability of ChatGPT to accurately detect and classify sentiment in various domains and text sources, including customer reviews, social media posts, and news articles. By harnessing the capabilities of ChatGPT's pre-trained language model, we seek to overcome the limitations associated with conventional supervised learning methods that heavily rely on extensive labeled data.
To tackle this issue, a comprehensive set of experiments and evaluations will be conducted using established benchmark datasets tailored for sentiment analysis tasks. The performance of ChatGPT will be compared against other cutting-edge sentiment analysis techniques, encompassing lexicon-based approaches and human labeling. By employing robust evaluation metrics such as accuracy, precision, recall, and F1-score, the effectiveness of ChatGPT in capturing the intricate aspects of sentiment across diverse domains will be thoroughly assessed.

\section{Experiments and Results}\label{ref:experimentaion}
This section provides information regarding the dataset's description, setup of the experiment, including the benchmarks, and techniques involved.
\subsection{Dataset}
For our experimentation, two datasets are used. The first one is from a soccer match between Manchester United and Liverpool.
The dataset is extensive and comes from a previous study where two independent annotators manually labeled tweets \cite{data}. This dataset contains 6,201 rows, of which 1,214 tweets contain emojis. Among the tweets, 2551 are positive, 2710 are negative, and 943 are neutral. The sports tweets dataset usually contains fewer words, and the use of emojis and abbreviations is common.
The second dataset used for experimentation is from Amazon reviews. It contains long text describing Amazon products \cite{dataset}. A random 2000 text example is taken as a subset from the dataset, with 1002 belonging to the positive category and 998 as negative.
The two datasets are different from each other based on the event as well as the length of an individual text.
\subsection{Setup}
For comparison, two lexical-based tools, VADER and TextBlob, are used. The evaluated ChatGPT variant is derived from GPT-3.5, specifically the gpt-3.5-turbo version, which is considered to be both highly capable and cost-effective. The prompt used has the same wording for both datasets. The numerical score has been extracted from the API response and hence used to label the sentiment.

\subsection{Result}
\subsubsection{Results: Soccer Tweets Dataset}

The accuracy of the VADER algorithm for the dataset was found to be 47\%, and the tweets containing emojis have an accuracy of 46\%. The accuracy of TextBlob's algorithm for the whole dataset is 40\%, and the tweets containing emojis are 33\% accurate. In contrast, the chatGPT is 67\% accurate for both the overall tweet and the tweet containing emojis.
\begin{figure}[h!]
    \centering
    \includegraphics[width=\columnwidth]{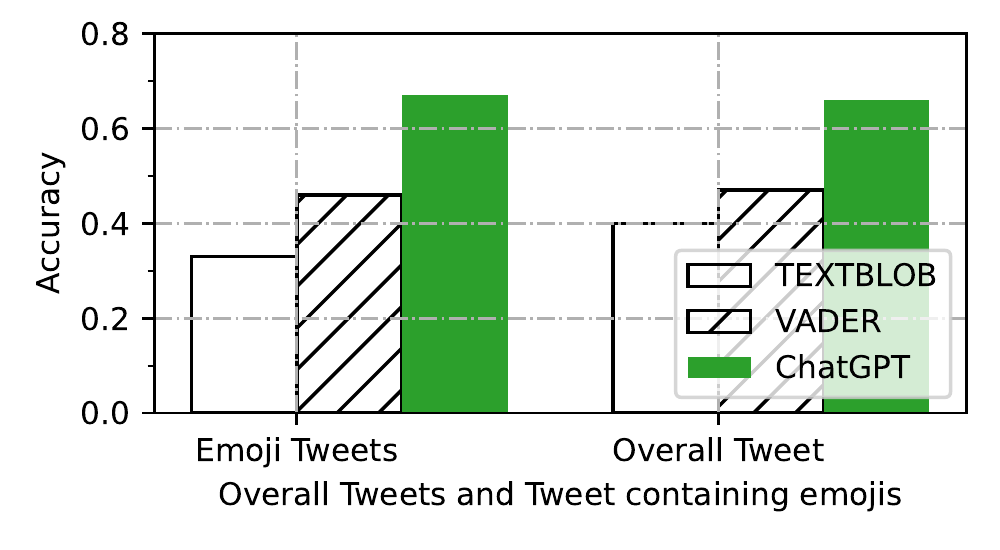}
    \caption{Accuracy of ChatGPT VS VADER and TextBlob}
    \label{fig:teams}
\end{figure}

\begin{table}[h!]
\caption{Result of the Soccer Tweets Dataset}

\begin{center}
\begin{tabular}{|p{1.2cm}|p{1.2cm}|p{1.5cm}|p{1.5cm}| p{1.3cm}|} 
\hline
Metric & Label & VADER's Algo & TextBlob Algo & ChatGPT \\
\hline
 \multirow{3}{4em}{Precision} & Positive & 0.51 & 0.51 & 0.75 \\ 
 & Neutral & 0.19 & 0.17 & 0.38\\
 & Negative & 0.74 & 0.70 & 0.75\\
 \hline

\multirow{3}{4em}{Recall} & Positive & 0.58 & 0.42 & 0.62 \\
 & Neutral & 0.37 & 0.48 & 0.59\\
 & Negative & 0.40 & 0.34 & 0.72\\

 \hline
\multirow{3}{4em}{F1-Score} & Positive & 0.55 & 0.46 & 0.68\\
 & Neutral & 0.25 & 0.25 & 0.46\\
 & Negative & 0.52 & 0.46 & 0.74\\

\hline

\end{tabular}
\end{center}
\label{table:resultTable}
\end{table}

\begin{table}[h!]
\caption{Result of Soccer Tweets Dataset containing emojis}

\begin{center}
\begin{tabular}{|p{1.2cm}|p{1.2cm}|p{1.5cm}|p{1.5cm}| p{1.3cm}|} 
\hline
Metric & Label & VADER's Algo & TextBlob Algo & ChatGPT \\
\hline
 \multirow{3}{4em}{Precision} & Positive & 0.50 & 0.45 & 0.71 \\ 
 & Neutral & 0.23 & 0.17 & 0.51\\
 & Negative & 0.58 & 0.64 & 0.71\\
 \hline

\multirow{3}{4em}{Recall} & Positive & 0.68 & 0.39 & 0.70 \\
 & Neutral & 0.22 & 0.41 & 0.61\\
 & Negative & 0.33 & 0.42 & 0.68\\

 \hline
\multirow{3}{4em}{F1-Score} & Positive & 0.58 & 0.42 & 0.70\\
 & Neutral & 0.23 & 0.24 & 0.58\\
 & Negative & 0.42 & 0.32 & 0.69\\

\hline

\end{tabular}
\end{center}
\label{table:resultTable}
\end{table}

\subsubsection{Results: Amazon Reviews Dataset}
The accuracy of the VADER algorithm for the Amazon reviews  dataset was found to be 69\%. The accuracy of TextBlob's algorithm for this dataset is 66\%, Whereas the chatGPT is 94\% accurate for the text it classified. ChatGPT could not classify 58 tweets as it termed them neutral or it needed more information.

\begin{figure}[h!]
    \centering
    \includegraphics[width=\columnwidth]{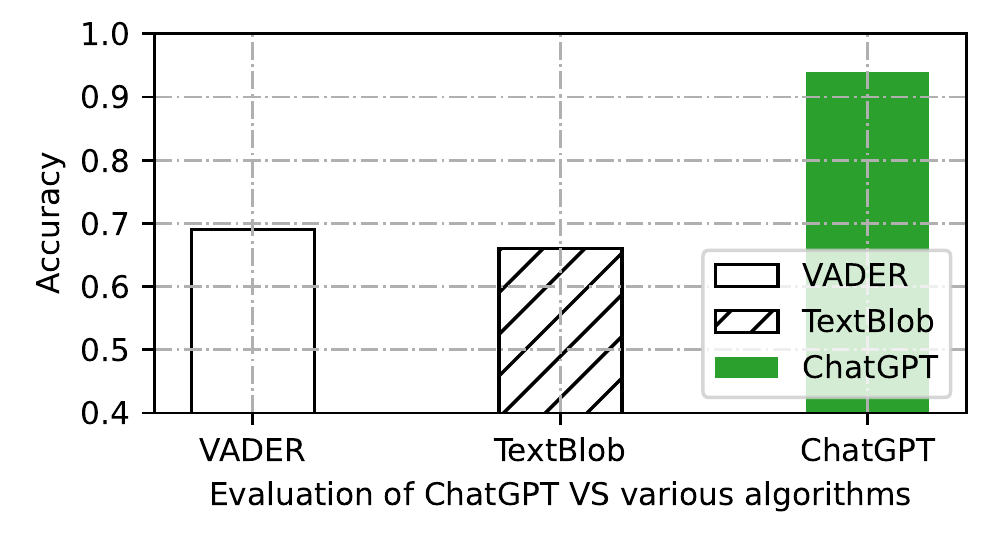}
    \caption{Accuracy of ChatGPT VS VADER and TextBlob}
    \label{fig:teams}
\end{figure}

\begin{table}[h!]
\caption{Result of Amazon Reviews Dataset}

\begin{center}
\begin{tabular}{|p{1.2cm}|p{1.2cm}|p{1.5cm}|p{1.5cm}| p{1.3cm}|} 
\hline
Metric & Label & VADER's Algo & TextBlob Algo & ChatGPT \\
\hline
 \multirow{3}{4em}{Precision} & Positive & 0.63 & 0.60 & 0.94 \\ 
 & Negative & 0.84 & 0.86 & 0.94\\
 \hline

\multirow{3}{4em}{Recall} & Positive & 0.91 & 0.94 & 0.95 \\
 & Negative & 0.46 & 0.38 & 0.94\\

 \hline
\multirow{3}{4em}{F1-Score} & Positive & 0.75 & 0.73 & 0.95\\
 & Negative & 0.60 & 0.52 & 0.94\\

\hline

\end{tabular}
\end{center}
\label{table:resultTable}
\end{table}

\section{Discussion}\label{sec: Discussion}
ChatGPT as a text annotation tool has shown promising potential and advantages. It offers a user-friendly and interactive interface, making it accessible to a wide range of users with varying levels of technical expertise. This ease of use allows non-experts to annotate text data without extensive training efficiently. Another advantage of using ChatGPT for text annotation is its adaptability to various annotation tasks and domains.
In terms of limitations, our study revealed that the sentiment value of a text depends on the prompt used for analysis. Furthermore, we observed instances where different sentiment scores were produced for the same prompt and tweet. One of the challenges is the potential bias in the model's responses, as it learns from the vast amount of data available on the internet, which can contain biased or controversial content. Careful consideration and monitoring are necessary to ensure that the annotations produced by ChatGPT are fair, unbiased, and aligned with the intended annotation guidelines. Additionally, the computational speed of ChatGPT is relatively slow compared to some other sentiment analysis tools. The processing time and cost associated with running the model may pose challenges, particularly for organizations with limited computational resources or budget constraints.

\section{Conclusion}\label{sec:Conclusion}

In this paper, we have examined the use of ChatGPT as a sentiment analysis tool. Sentiment analysis has become increasingly important in understanding customer feedback and opinions in today's digital world. While supervised learning algorithms have been popularly used for this task, they require human-annotated text to train the classifier, which can be time-consuming and expensive. To address this challenge, lexicon-based tools have been used, and in recent times, zero-shot models have gained popularity. This study evaluated ChatGPT as a tool for sentiment analysis labeling on two distinct datasets: sports tweets and Amazon reviews. It outperforms the lexical-based tools by 20\% and 25\% on these two datasets. ChatGPT shows promise as a tool for text labeling for  various  sentiment analysis events and tasks. 
As for future directions, there is a need to optimize and parallelize the computational tasks of ChatGPT to improve its speed and efficiency. Furthermore, an examination of sarcasm and irony detection by ChatGPT could be evaluated. 
\ifCLASSOPTIONcompsoc
  \section*{Acknowledgments}
\else
  \section*{Acknowledgment}
\fi
This work was initiated by the College of Science and Engineering at HBKU.

\ifCLASSOPTIONcaptionsoff
  \newpage
\fi

\bibliographystyle{IEEEtran}
\bibliography{ref}

\end{document}